\documentclass[manuscript,screen]{acmart}
\AtBeginDocument{%
  \providecommand\BibTeX{{%
    \normalfont B\kern-0.5em{\scshape i\kern-0.25em b}\kern-0.8em\TeX}}}

\setcopyright{acmcopyright}
\copyrightyear{2020}
\acmYear{2020}
\acmDOI{10.1145/1122445.1122456}

\acmConference[IoT4Emergency]{1st international workshop on Internet of Things for Emergency Management}{October 6, 2020}{Malm\"o, Sweden}
\acmBooktitle{1st International Workshop on Internet of Things for Emergency Management, October 6, 2020, Malm\"o, Sweden}
\acmPrice{15.00}
\acmISBN{978-1-4503-XXXX-X/18/06}

\begin{document}

\title[Enabling Image Recognition on Constrained Devices]{Enabling Image Recognition on Constrained Devices Using Neural Network Pruning and a CycleGAN}

\author{August Lidfeldt}
\authornote{These authors contributed equally to this research.}
\affiliation{%
  \institution{Dept. of Computer Science, Lund University}
  \city{Lund}
  \country{Sweden}}
\email{august.lidfeldt@gmail.com}

\author{Daniel Isaksson}
\authornotemark[1]
\affiliation{%
  \institution{Dept. of Computer Science, Lund University}
  \city{Lund}
  \country{Sweden}}
\email{daniel.g.isaksson@gmail.com}

\author{Ludwig Hedlund}
\authornotemark[1]
\affiliation{%
  \institution{Dept. of Computer Science, Lund University}
  \city{Lund}
  \country{Sweden}}
\email{ludwighedlund@outlook.com}

\author{Simon Åberg}
\authornotemark[1]
\affiliation{%
  \institution{Dept. of Computer Science, Lund University}
  \city{Lund}
  \country{Sweden}}
\email{simon.aberg95@gmail.com}

\author{Markus Borg}
\affiliation{%
  \institution{RISE Research Institutes of Sweden}
  \city{Lund}
  \country{Sweden}}
\email{markus.borg@ri.se}

\author{Erik Larsson}
\affiliation{%
  \institution{Dept. of Electro and Information Technology, Lund University}
  \city{Lund}
  \country{Sweden}}
\email{erik.larsson@eit.lth.se}

\renewcommand{\shortauthors}{Lidfeldt \textit{et al.}}


\begin{abstract}
Smart cameras are increasingly used in surveillance solutions in public spaces. Contemporary computer vision applications can be used to recognize events that require intervention by emergency services. Smart cameras can be mounted in locations where citizens feel particularly unsafe, e.g., pathways and underpasses with a history of incidents. One promising approach for smart cameras is edge AI, i.e., deploying AI technology on IoT devices. However, implementing resource-demanding technology such as image recognition using deep neural networks (DNN) on constrained devices is a substantial challenge. In this paper, we explore two approaches to reduce the need for compute in contemporary image recognition in an underpass. First, we showcase successful neural network pruning, i.e., we retain comparable classification accuracy with only 1.1\% of the neurons remaining from the state-of-the-art DNN architecture. Second, we demonstrate how a CycleGAN can be used to transform out-of-distribution images to the operational design domain. We posit that both pruning and CycleGANs are promising enablers for efficient edge AI in smart cameras.

\end{abstract}




\maketitle

\section{Introduction}
According to a study conducted in 2019 by The Swedish National Council for Crime Prevention, 28\% of Swedes felt unsafe walking outside in their own neighbourhood at night. This number marks an increase from 21\% in 2013 and is indicative of the larger trend of lower perceived safety. The same study also reported an increasing concern about crime in society from 28\% in 2013 to 43\% in 2019~\cite{molin2019swedish}. The trend leads to the larger question of how to create safe societies. 

%One can expect to see a multitude of different measures in the coming years, aimed at curbing this trend. The appropriateness of the chosen interventions will most likely dictate the outcome. 
A measure that often is proposed is strategic placement of cameras in public spaces~\cite{Guo2018CrowdAICS,Choi2020EdgeCS}. Although cameras are increasingly used in surveillance solutions, operators typically are required to detect and classify incidents. AI and machine learning (ML) enable smart cameras~\cite{liu2013intelligent} that allow automatic recognition of events that require intervention by emergency services.  Large scale camera deployment increase the bandwidth requirements, leading to a need to distribute computation to the cameras themselves. Edge AI is based on the idea of decentralized computational platforms, where AI technology such as image recognition is incorporated directly in IoT devices~\cite{shi2016edge}. 

Image recognition on the constrained edge devices introduces fundamental trade-offs between performance and efficiency. The pursuit of high accuracy image recognition has lead to ever-growing deep neural network (DNN) architectures. State-of-the-art DNN architectures contain trainable parameters in the magnitude of hundreds of millions, which requires considerable computational power and energy. To mitigate the issue, several studies have investigated neural network pruning, i.e., decreasing the size of DNNs by reducing the number of trainable parameters while trying to retain model accuracy~\cite{han2015deep}.

Robustness is another essential quality attribute in image recognition, especially for critical emergency response applications. For a trained DNN, robustness involves handling perturbations or input data that is does not closely resemble the training data -- input referred to as being out-of-distribution (OOD)~\cite{henriksson2019performance}. In the context of smart cameras, perturbations might include dirty or vandalized camera domes. An example that could lead to OOD input would be deployment of the camera in an environment that does not reflect the training data, e.g., due to differences in illumination. Inspired by work in the automotive domain~\cite{porav2019can}, we propose using a CycleGAN to perform style transfers of OOD input.

In this paper, we study a motion-activated network camera mounted in an underpass located in Helsingborg, Sweden. We use the camera input to train various DNNs for image recognition, using the well-known VGG16~\cite{VGG14} as the baseline. In this preliminary work, we discuss an application of multi-class classification, i.e., detecting the presence of pedestrians, dog walkers, and bicyclists. Again inspired by automotive engineering, we specify the operational design domain (ODD)~\cite{gyllenhammar2020towards} of our classification model to cover daytime conditions. Two research questions (RQ) guide us:

\begin{itemize}
    \item[RQ1] How does neural network pruning of a state-of-the-art DNN architecture affect the classification accuracy?
    \item[RQ2] How can a CycleGAN be used to transfer OOD input to the ODD of an image recognition application?
\end{itemize}

Our results show that substantial pruning of VGG16 is possible in our case under study. Given the ODD, i.e., homogeneous daytime conditions in the underpass, we obtain a classification accuracy above 90\% despite pruning the DNN to contain only 1.1\% of the trainable parameters. Second, as a proof-of-concept, we report how style transfers from nighttime to daytime conditions improves the classification accuracy of OOD input images -- the CycleGAN might thus extend the ODD of the smart camera by performing classification beyond its underlying training data. We hypothesize that style transfers can be an effective and efficient way to enable smart cameras to operate in environments for which their embedded DNNs were not trained.

The rest of this paper is organized as follows. Section~\ref{sec:bg} introduces fundamental concepts related to image classification and CycleGANs. In Section~\ref{sec:method}, we present the overall ML workflow and two experiments that address the RQs. Section~\ref{sec:res} presents the results and the discussion follows in Section~\ref{sec:disc}. Section~\ref{sec:threats} reports the main threats to validity. Finally, Section~\ref{sec:conc} concludes the paper and outlines directions for future work.

\section{Background} \label{sec:bg}
Thanks to DNNs and massive datasets, image recognition has been reported to outperform humans on specific tasks in the last decade. A key component in this development is convolutional neural networks (CNN)~\cite[pp. 321-362]{goodfellow2016deep}. The main purpose of CNNs is to extract key features from images in a computationally efficient way. CNNs are generally made up of several different types of layers with two of the most important being fully connected layers and convolutional layers. 

A large computational expense in traditional neural networks stems from the heavy reliance on fully connected layers, where each neuron is connected to all the neurons in the previous layer. CNNs combine fully connected layers with convolutional layers in which each neuron only is connected to a small region of the input volume and thereby greatly reduces the number of trainable parameters and the computational cost. Convolutional layers iterate sequentially over sections of the image and produces an edge feature representation derived from contrasts in light and color. From this the network will learn which specific features at a given spatial position of the input that should trigger an activation.

In this work, we use VGG16 as the baseline DNN architecture -- a well-known convolutional DNN that has obtained accurate results in recognized competitions, e.g., the ImageNet Large Scale Visual Recognition Competition\cite{VGG14}.

Generative Adversarial Networks (GAN) consist of two competing models, a generator and a discriminator~\cite[pp. 690-693]{goodfellow2016deep}. The generator generates fake samples of data, images in our case, while the discriminator’s purpose is to distinguish if the samples come from the original dataset or are generated by the generator. These models get updated by an \textit{adversarial loss} where the generator tries to maximize the discriminator probability of incorrectly labeling the generated sample as fake. The discriminator on the other hand tries to minimize the same object function. When these models are successfully trained against each other the generator learns to produce random samples indistinguishable from the original training collection.

Zhu \textit{et al}. proposed CycleGANs, an architecture that combines two GANs~\cite{Zhu_2017_ICCV}. CycleGANs have been trained to transform images between domains while preserving the images’ specific features. A CycleGAN is trained on unlabeled datasets from the chosen domains, i.e., unpaired image to image translation. For example, if camera images in rainy conditions are underrepresented in the training set, existing images from the sunny domain can be transformed to the rainy domain as an approach to data augmentation~\cite{huang2018auggan}. Moreover, CycleGANs can be used during operation to transform real-time input data that does not resemble the training data to images that are within the ODD~\cite{anoosheh2019night}.

A CycleGAN is, as the original GAN, optimized by an \textit{adversarial loss} but the CycleGANs generators have three additional losses, \textit{identity loss}, \textit{forward-} and \textit{backward cycle consistency loss}. The \textit{identity loss} is calculated by the difference between input and output when the input for the generator is in the same domain as the target domain. The \textit{forward-} and \textit{backward cycle consistency loss} is important to ensure that the output keeps the specific features of the input image and not only generates a sample that resembles the target domain. These losses are calculated when you combine the two different generators in a cycle which put the input and output in the same domain and then calculate the difference between input and output.

\section{Method} \label{sec:method}
This section describes the training data, the machine learning process, and the experimental setup. A complete replication package is available on GitHub\footnote{https://github.com/luuddan/EITN35}.

\subsection{End-to-End Machine Learning Process}
\begin{figure*}
  \centering
  \includegraphics[width=\linewidth]{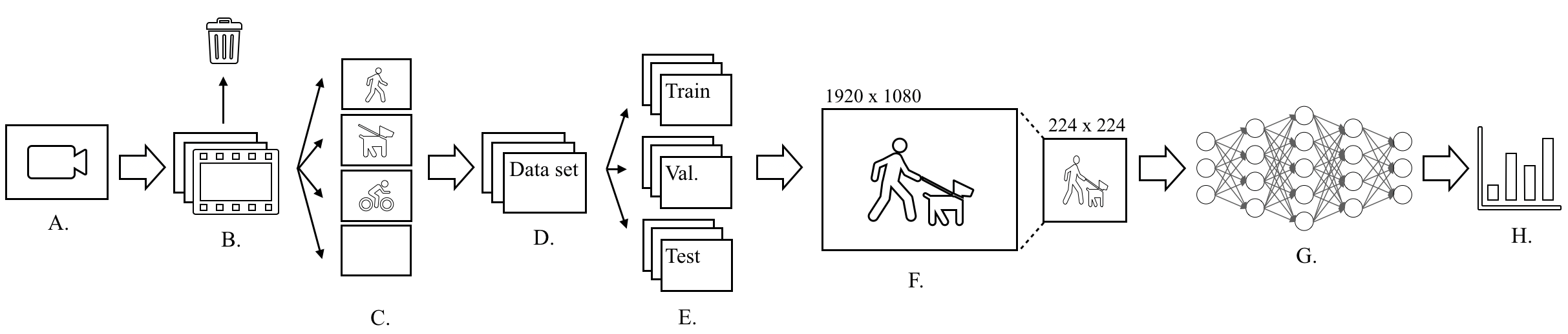}
    \caption{Overview of the end-to-end machine learning process.}
  \label{fig:endtoend}
\end{figure*}

\begin{figure}
  \centering
  \includegraphics[width=\linewidth]{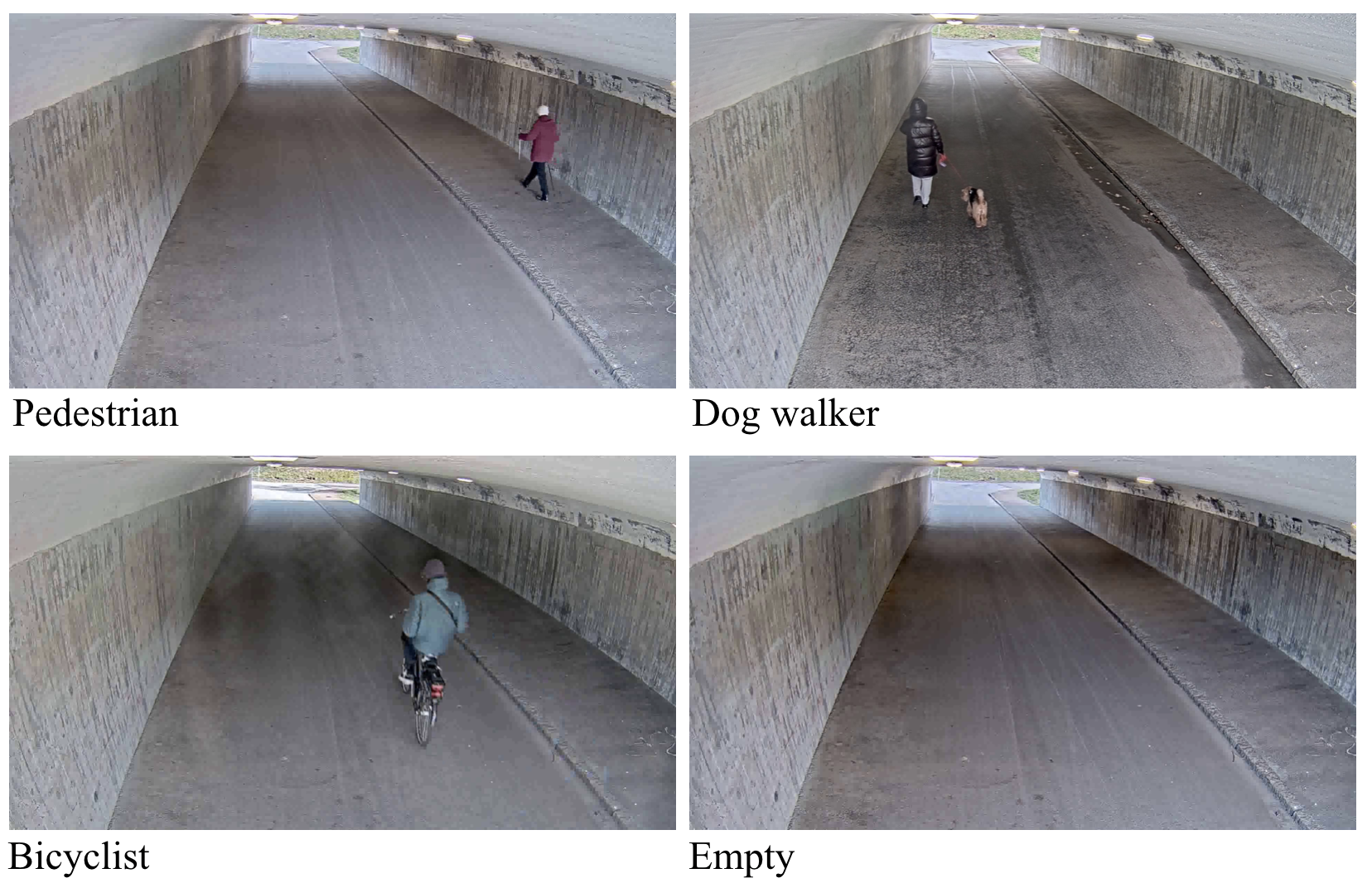}
  \caption{Example images of each class.}
  \Description{Example images of each class}
  \label{fig:examples}
\end{figure}

Figure~\ref{fig:endtoend} shows how data were collected and processed. First, we collected 90 minutes of high resolution video clips from the underpass (A) with an average video clip length of 52 seconds. Note that all recordings were action-triggered, i.e., all clips contain activity in the underpass. All raw videos were recorded during five days in the spring of 2020 with a fixed camera angle as presented in Figure~\ref{fig:examples}.

We split the video clips into individual frames (B) at a frame rate of one per second. After step B), we had a dataset containing 5,978 single images. The first four authors manually annotated the images (C) with one of the four labels 1) pedestrian, 2) dog walker, 3) bicyclist, or 4) empty. Images that did not fit into any of these classes or that contained more than one instance of a label were removed in this step. Images that did not match our quality criteria also were excluded, e.g., objects in the far end of the underpass and objects only partially visible. After the labelling step, the final dataset (D) consisted of 180 dog walkers, 904 pedestrians, and 253 bicyclists. To maintain a useful class distribution, we kept only 904 images with the empty label.

In line with standard practice in machine learning, we randomly split the dataset (E) into a training set (64 \%), a validation set (16 \%), and a test set (20 \%). As a final step before using the images as input to the DNN, we downscaled them (F) from a resolution of 1920x1080 to 224x224 pixels. 

We trained a DNN for the multi-class classification task (G) using VGG16 as the baseline DNN architecture (G). VGG16 is composed of 16 layers with trainable parameters, whereof the first 13 are convolutional (followed by max-pooling layers) and the last three are fully-connected. The total number of trainable parameters in VGG16 is 134 million. We use early stopping to mitigate overfitting when training all models.

\subsection{Neural Network Pruning (RQ1)}
Experiment A investigates what effect the number of trainable parameters the DNN architecture has on the classification accuracy. Our baseline architecture (Arch11) is structurally identical to the VGG16 model. We then created 10 additional architectures (Arch1-10) by iteratively reducing the number of parameters by (roughly) 50\% in each step. The reduction approach consisted of alternating between removal of convolutional layers, reducing the number of filters in the convolutional layers, and shrinking the dense layers. Table~\ref{tab:archs} presents an overview of the 11 architectures. Note that after training classification models using these architectures, we instead refer to them as models (M1-M11).

\begin{table*}
  \caption{Characteristics of the 11 investigated DNN architectures. Arch11 is the VGG16 baseline.}
  \label{tab:archs}
  \begin{tabular}{cccccccccccl}
    \toprule
    Arch1&Arch2&Arch3&Arch4&Arch5&Arch6&Arch7&Arch8&Arch9&Arch10 & Arch11\\
    \midrule
     \ 0.1 M & 0.2 M & 0.4 M & 0.8 M & 1.5 M & 3 M& 6 M & 13 M &35 M & 67 M &134 M & \#Parameters\\
     \ 2 & 3 & 3 & 4 & 4 & 4 & 4 & 8 & 10 & 11 & 13 &
     \#Convolutional layers\\
     \ 8 & 16 & 32 & 128 & 128 & 256 & 256 & 512 & 1,024 & 2x2,048 & 2x4,096 & \#Dense layers\\
  \bottomrule
\end{tabular}
\end{table*}

Experiment A uses a full factorial design with two independent variables with discrete values. First, the \textit{DNN architecture} is varied by training classification models using the 11 DNN architectures listed in Table~\ref{tab:archs}. Second, we varied the \textit{amount of training data} by creating subsets containing 25\%, 50\%, and 75\% of the final dataset. The subsets were created through random stratified sampling, i.e., the dataset was split into new training, validation, and test sets while retaining the original class distributions.

For the most promising DNN architecture from experiment A, we performed hyperparameter tuning using grid search, i.e., we evaluated a manually specified subset of the hyperparameter space of the learning algorithm. We used the one-factor-at-a-time method for the tuning, i.e., we did not investigate interaction effects~\cite{borg2016tuner}. 

\subsection{CycleGAN Transformation to ODD (RQ2)}
We designed Experiment B to act as a proof-of-concept for the approach to use a CycleGAN to transform OOD input to the ODD. In our case, we explore whether a CycleGAN can transform input images from the nighttime domain to the daytime domain. 

We used an open source implementation\footnote{https://machinelearningmastery.com/cyclegan-tutorial-with-keras/} of the CycleGAN architecture proposed by Zhu \textit{et al.}~\cite{Zhu_2017_ICCV}. The architecture consists of two discriminator models and two generator models. The discriminator models consist of 5 convolutional layers. For input, the discriminator models take both real and generated images and outputs a binary value reflecting whether the input was real or fake. We used a mean square error loss as the loss function for the discriminator model. 

The generator models are using an encoder-decoder approach. Input images are downsampled to extract the features and then upsampled into a new image in the new domain based on the extracted features. 

The encoder consists of three convolutional layers followed by a section of six ResNet blocks which are used in deep neural networks for convergence while avoiding exploding or vanishing gradients\cite{DBLP:journals/corr/HeZRS15}. Next, the decoder follows, consisting of two transpose convolutional layers, i.e. convolutional layers upscaling the resolution as opposed to normal convolutional layers which downscale it. Lastly a final normal convolutional layer follows.

%\begin{figure}
%\includegraphics[width=8cm]{CycleConsistencyLoss.png}
%\caption{A CycleGAN mapping from the domain X to \^{Y} and then reconstructing it back to \^{X}. The numbers show the four losses used to optimize the process.}
%\end{figure}

%The generator models are optimized by four different calculated losses. First, the \textit{adversarial loss} where the generator tries to generate images with a low probability that the discriminator guesses it being fake. The generator also gets updated by an \textit{identity loss} which is calculated through the difference between input and output when the input for the generator is in the same domain as the target domain. The last losses are \textit{forward and backward cycle consistency loss}. These losses are calculated when you combine the two different generators in a cycle which put the input and output in the same domain and then calculate the difference between input and output.%, as shown in Figure 3.

\begin{figure}
    \centering
    \includegraphics[width=1.0\linewidth]{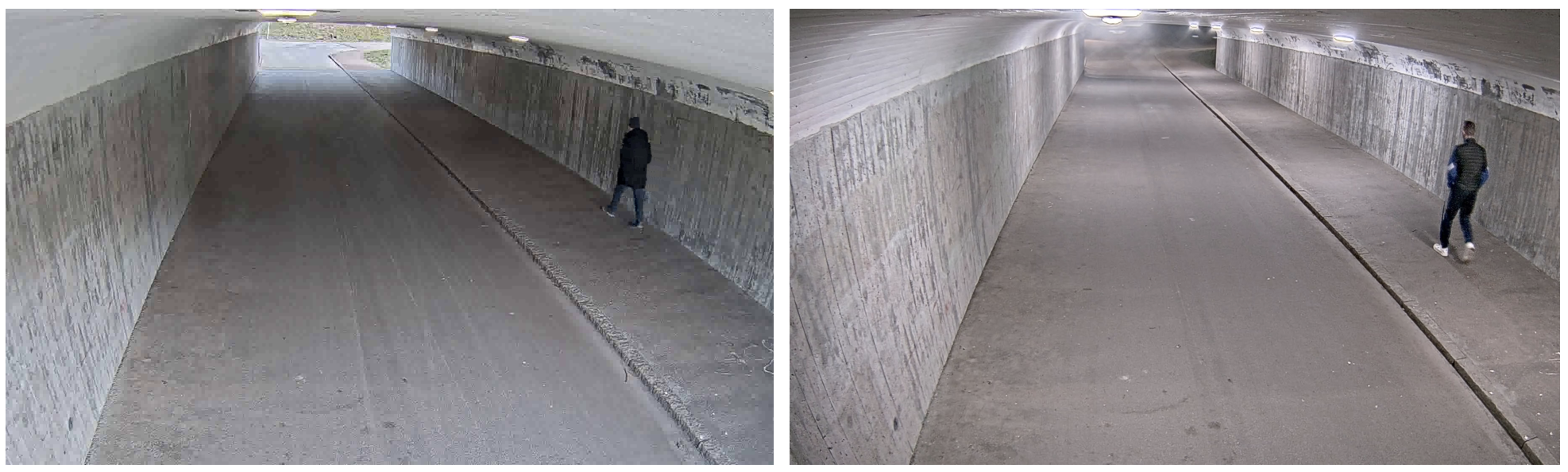} 
    \caption{Illumination in the daytime domain (left) and the nighttime domain (right)}
    \label{fig:domainexamples}
\end{figure}

We trained the CycleGAN using a dataset containing 1,128 images from the underpass with objects close to the camera, i.e., a sample from step B in Figure~\ref{fig:endtoend}. The dataset contained an equal share of images from the daytime domain and the nighttime domain. Figure~\ref{fig:domainexamples} illustrates how the illumination differs in the two domains.

\begin{figure}
\includegraphics[width=0.8\linewidth]{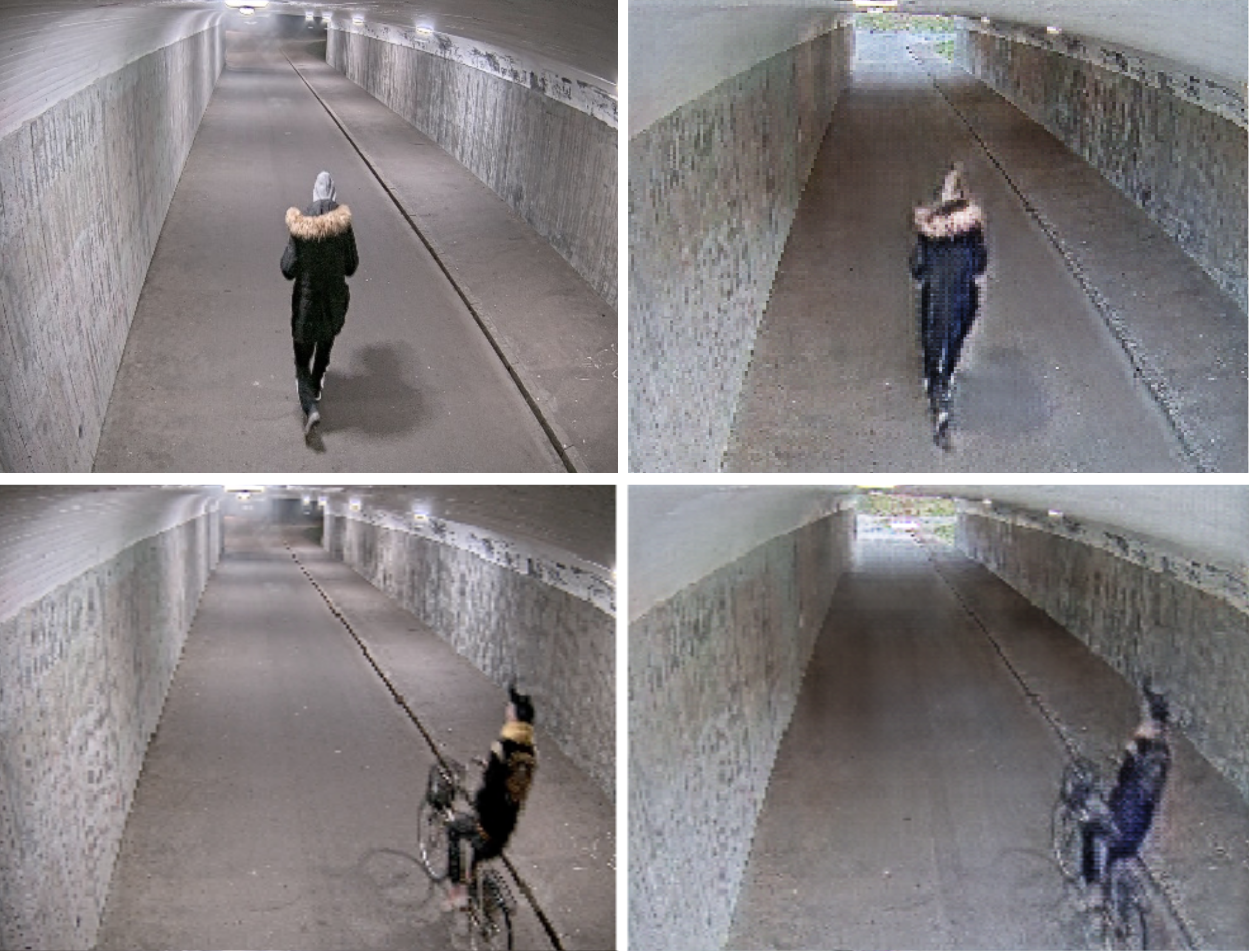}
\caption{Images transformed from nighttime (left) to daytime (right).}
\label{fig:transformations}
\end{figure}

Experiment B constituted an evaluation of M5 from Experiment A, trained on 100\% of the training data (step D in Figure~\ref{fig:endtoend}). Table~\ref{tab:gantests} describes the three new test sets we created for this evaluation, containing images with pedestrians (PedSet), bicyclists (BikeSet), and empty underpass (EmpSet), respectively. Each test set contained a combination of randomly sampled images from the CycleGAN dataset from the daytime and nighttime domains. Furthermore, we used the trained CycleGAN to transform the nighttime images to the daytime domain (cf. Night2Day in Table~\ref{tab:gantests}). Two examples of transformed images are presented in Figure~\ref{fig:transformations}.

\begin{table}[]
\caption{Distribution of images in the three test sets.}
\label{tab:gantests}
\begin{tabular}{ll|c|c|c|}
\cline{3-5}
                                        &                & \multicolumn{3}{c|}{\textbf{Domain}}                                                                                      \\ \hline
\multicolumn{1}{|l|}{\textbf{Test set}} & \textbf{Label} & \multicolumn{1}{l|}{\textbf{Day}} & \multicolumn{1}{l|}{\textbf{Night}} & \multicolumn{1}{l|}{\textbf{Night2Day}} \\ \hline
\multicolumn{1}{|l|}{PedSet}            & Pedestrian     & 180                                   & 106                                     & 106                                     \\ \hline
\multicolumn{1}{|l|}{BikeSet}           & Bicyclist      & 50                                    & 60                                      & 60                                      \\ \hline
\multicolumn{1}{|l|}{EmpSet}            & Empty          & 180                                   & 106                                     & 106                                     \\ \hline
\end{tabular}
\end{table}

\section{Results} \label{sec:res}
This section presents experimental results concerning the two RQs.

\subsection{Neural Network Pruning (RQ1)} \label{sec:rq1res}
\begin{figure}[h]
  \centering
  \includegraphics[width=0.8\linewidth]{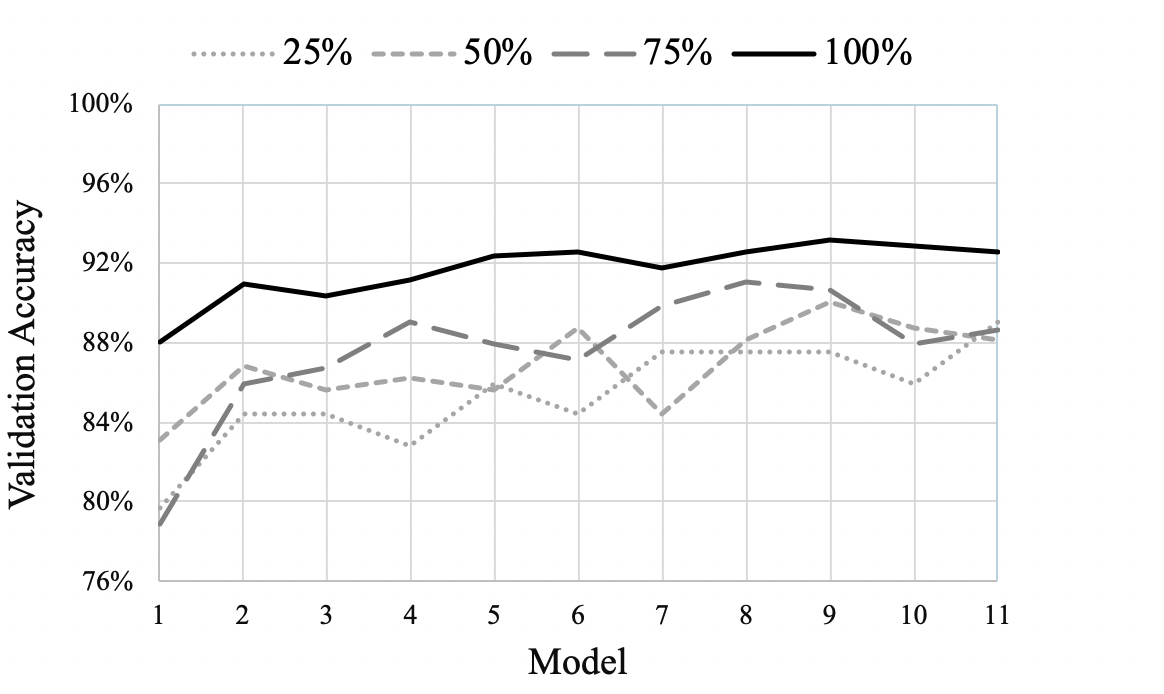}
  \caption{Validation accuracy per model. The four lines show the size of the datasets.}
  \label{fig:valacc}
\end{figure}

Figure~\ref{fig:valacc} shows the classification accuracy on the validation set from Experiment A. The X-axis shows classification models trained according to the architectures listed in Table~\ref{tab:archs}. The models shall be considered on an ordinal scale, with increasing numbers of trainable parameters toward the right. 

The four lines represent validation accuracy for different dataset sizes. The solid black line, corresponding to 100\% of the dataset, displays as expected the best results. As all lines show increasing trends, the results suggest that more complex DNN architectures result in more accurate object recognition in the underpass. 

On the other hand, training models with more data influences the accuracy more than having more complex DNN architectures. The second smallest model (M2) using 100\% of the dataset outperforms all of the more complex models using 25\%-75\% of the data (except M8 that obtained almost the same validation accuracy). This observation is in line with previous work on object recognition~\cite{VGG14,luo18}, and demonstrates the potential of pruning DNN architectures while retaining acceptable accuracy.

\begin{figure}[h]
  \centering
  \includegraphics[width=0.8\linewidth]{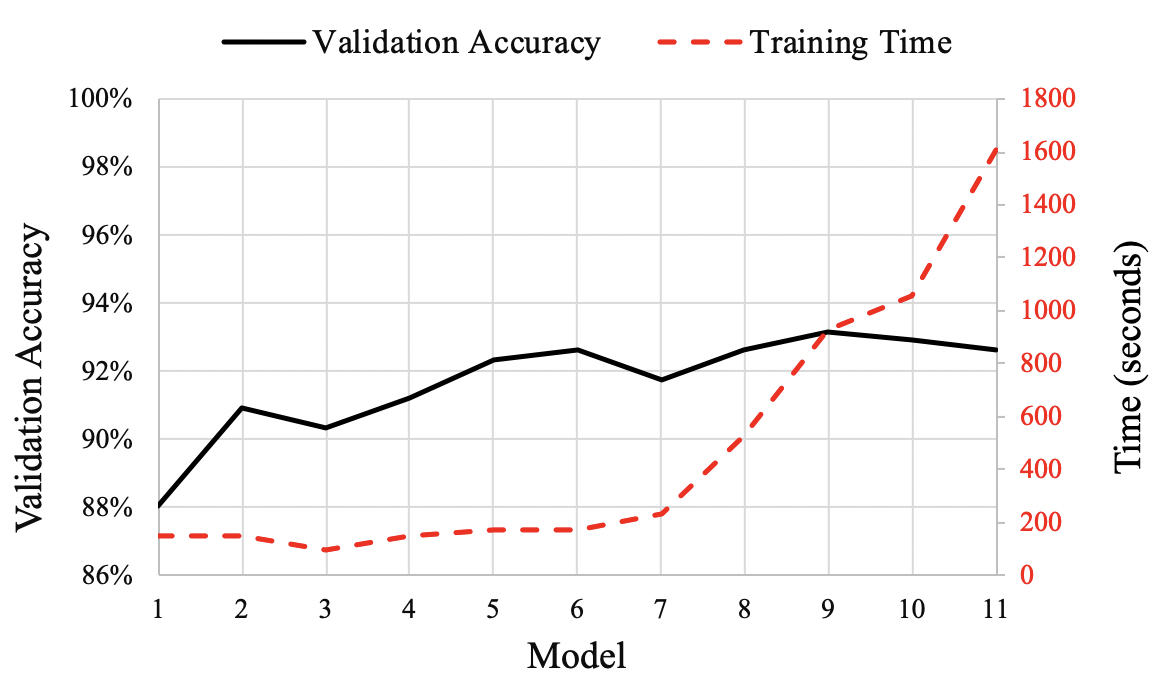}
  \caption{Validation accuracy per model with respective training time}
  \label{fig:time}
\end{figure}

Figure~\ref{fig:time} depicts the training time for the different models. The results correspond to training using the complete dataset until early stopping (avg. \#epochs=115, SD=16). All measurements are reported in seconds, as shown on the Y-axis to the right. The results show that the training times drastically increase with more complex DNN architectures, i.e., from M8 containing 65M trainable parameters. 

Based on the results from Experiment A, we selected M5 for further development and evaluation. Our rationale was that M5 was the smallest model that performed within 1\% of the baseline VGG16 architecture (M11) -- M5 has 1.5M trainable parameters, corresponding to 1.1\% of M11. Table~\ref{tab:hyperparams} lists the hyperparameters that we tuned for M5, the values we evaluated, and the final settings we used. Our final results after hyperparameter tuning on the training set and the validation set were 98.8\% and 93.5\%, respectively.

\begin{table}
\caption{Hyperparameters evaluated during tuning and the final setting in bold font.}
\begin{tabular}{ll}
Hyperparameter            & Values \\
\hline
Learning rate             & 5E-5, 1E-4, 5E-4, \textbf{0.001}, 0.005, 0.01         \\
Dropout rate              & 0, 0.1, ... 0.45, \textbf{0.5}, 0.55, ... 0.65         \\
%& 0, 0.1, 0.2, 0.25, 0.3, 0.35, 0.4, 0.45, \textbf{0.5}, 0.55, 0.6, 0.65         \\
L2 Regularization rate    & \textbf{0}, 1E-4,	5E-4,    0.001,	0.005,	0.01,	0.1          \\   
\end{tabular}
\label{tab:hyperparams}
\end{table}

Table~\ref{tab:testset} shows the classification accuracy on the test set. M5 obtained an overall accuracy of 91.0\%. Looking at the individual classes, we note the least accurate results for the dog walker class (57.1\%) whereas the empty underpass was no problem for the classifier. 

\begin{table}
  \caption{M5 Classification Accuracy on the test set.}
  \label{tab:testset}
  \begin{tabular}{cc}
    \toprule
    Class & Accuracy\\
    \midrule
     \ \textbf{Total} & 91.0\%\\
    \ \textbf{Empty} & 100.0\%\\
    \ \textbf{Pedestrian} &92.0\%\\
     \ \textbf{Dog walker} &57.1\%\\
       \ \textbf{Bicyclist} &80.0\%\\
  \bottomrule
\end{tabular}
\end{table}

\begin{figure}
  \centering
  \includegraphics[width=0.7\linewidth]{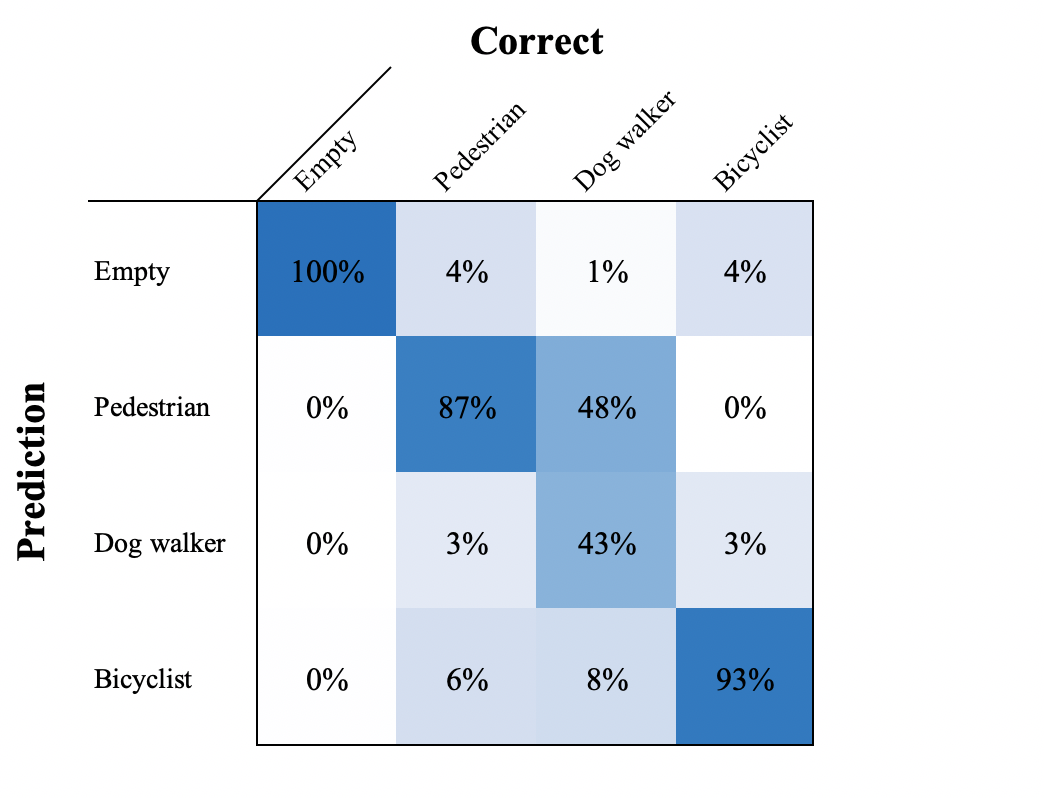}
  \caption{Confusion matrix for M5}
  \label{fig:conf_evolve}
\end{figure}

Figure~\ref{fig:conf_evolve} presents a confusion matrix, enabling further examination of erroneous predictions. The confusion matrix shows that for input of the class dog walker, the classifier could not properly distinguish between pedestrians (48\%) and the correct class (43\%). We report two possible explanations. First, the proportion of dog walkers in the training set was low (8\%). Second, the differences between some images of pedestrians and dog walkers are indeed minor, e.g., small dog breeds on a leash.

\subsection{CycleGAN Transformation to ODD (RQ2)} \label{sec:rq2res}
\begin{figure}
  \centering
  \includegraphics[width=0.75\linewidth]{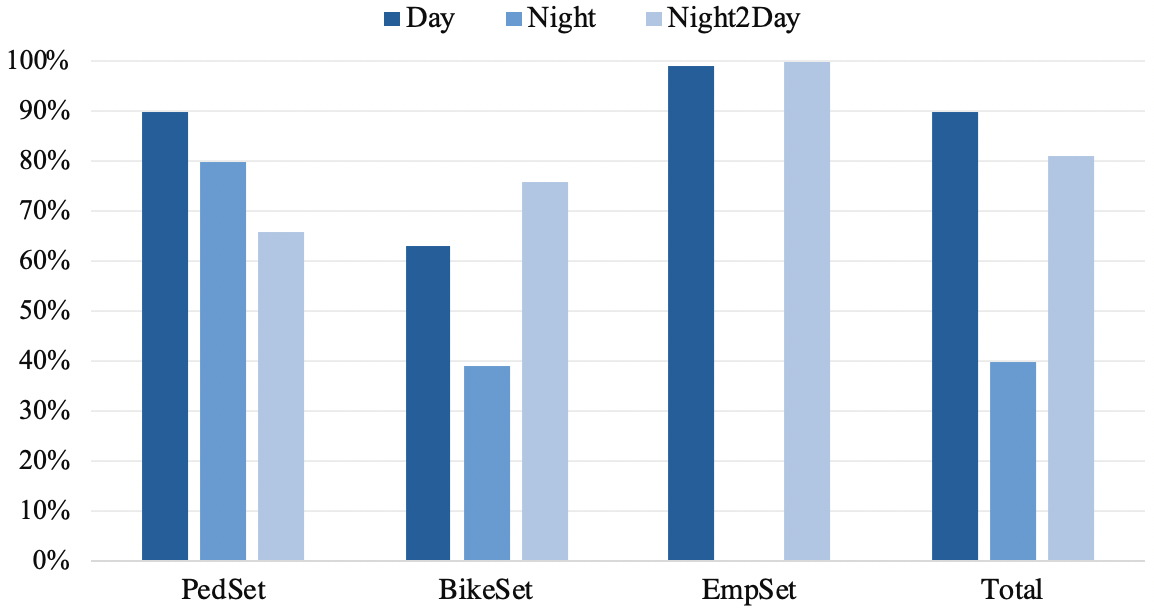} 
  \caption{Classification accuracy for M5 on EmpSet, PedSet, BikeSet, and results for the combined test sets (Total).}
  \label{fig:cycleGANres}
\end{figure}

Figure~\ref{fig:cycleGANres} reports the classification accuracy of M5, after hyperparameter tuning, on PedSet, BikeSet, and EmpSet as well as the overall result. For EmpSet, the different illuminations in the daytime and nighttime domains yield completely opposite results. All predictions in daytime are correct, but none in nighttime. However, after the Night2Day transformation back to the ODD, M5 again obtains a perfect result.

The classification results for PedSet and BikeSet are contrasting. For PedSet, M5 obtains 90\% and 80\% classification accuracy for the daytime and nighttime domains, respectively. The accuracy for the images transformed to the ODD using Night2Day was only 65\%. The results for BikeSet, on the other hand, were orthogonal. M5 obtains 60\% accuracy for the daytime domain and 35\% for nighttime, but the Night2Day transformation enables a substantial improvement -- 75\% of the bicyclists are correctly classified. We consider this a promising proof-of-concept for ODD extension using a CycleGAN, i.e., input images that do not resemble the training data can be be transformed to the ODD.

\begin{figure}
  \centering
  \includegraphics[width=0.9\linewidth]{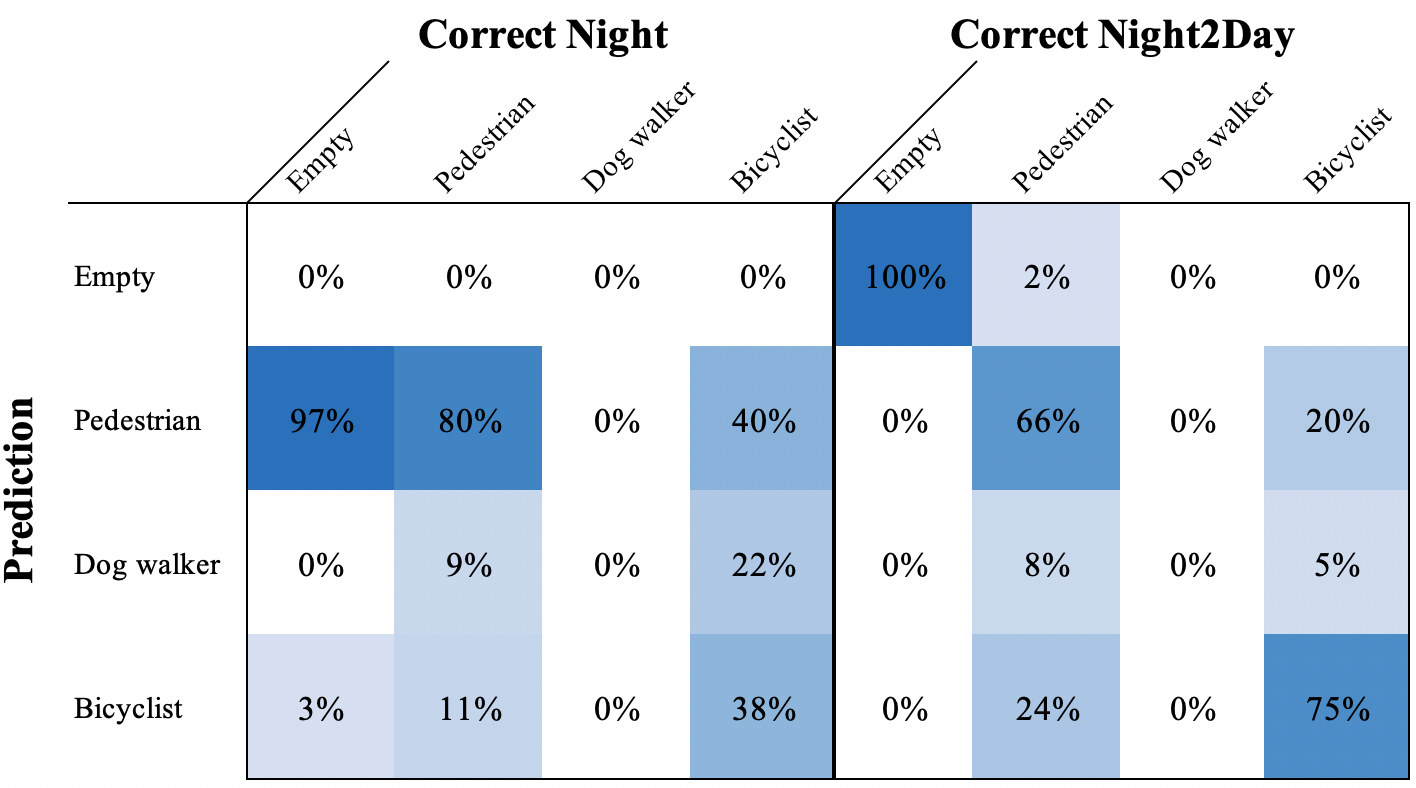}%{conf_matrix_ASSESS_16_v2.png}
  \caption{Confusion matrices for M5 on images in the nighttime domain (left) and images transformed from night2day (right).}
  \label{fig:conf_assess_night2day}
\end{figure}

Figure~\ref{fig:conf_assess_night2day} presents confusion matrices for the nighttime domain and the images transformed from night to day, respectively. For the nighttime domain, we notice that an empty underpass in most cases (97\%) resulted in the pedestrian label, i.e., the M5 classifier identified features suggestion people in the underpass background. Furthermore, we highlight that M5 predicted input of bicyclists in the nighttime domain as pedestrians, dog walkers, and bicyclists rather arbitrarily. Figure~\ref{fig:conf_assess_night2day} shows how the Night2Day transformation mitigated this, as 75\% of the bicyclists were correctly classified. Moreover, after the Night2Day transformation, M5 identified no pedestrians in input images containing an empty underpass. 

%\begin{figure}
%  \centering
%  \includegraphics[width=0.8\linewidth]{conf_matrix_ASSESS_15_v2.png}
%  \caption{Confusion matrix for M5 on the nighttime domain.}
%  \label{fig:conf_assess_night}
%\end{figure}

\section{Discussion} \label{sec:disc}
Enabling dependable image recognition on edge devices is an important topic to realize IoT solutions for emergency management. Our study demonstrates two approaches that make applications of edge AI practically feasible on constrained devices in a controlled environment such as an underpass.

Neural network pruning can be used to substantially reduce the size of a DNN model for image recognition. Reduced DNNs lead to reduced needs for computation as well as limited energy consumption~\cite{Hao19}. In line with previous work~\cite{Lucero17,pruning20}, we found that substantially smaller DNNs can perform comparably in terms of classification accuracy. We hypothesize that the pruning worked particularly well in our case due the relatively low complexity in the recognition task, a low number of classes, and the static image background. VGG16, on the other hand, was developed to classify 1,000 labels in arbitrary input. For our specific application, deploying VGG16 on an edge device would have constituted considerable over-engineering.

Changing the DNN size between Arch11's maximum of 134 million trainable parameters down to 0.1 million for Arch1 did have an effect on our validation accuracy, but not as large of a change as one might expect. Decrease the number of trainable parameters by three magnitudes only resulted in a validation accuracy drop of 4.5\%. While our results show the potential of pruning, finding the appropriate balance between DNN size and accuracy is truly application specific -- conflicting quality requirements must be managed.

Our results also highlight the trade-off between classification accuracy and training time (cf. Figure~\ref{fig:valacc}). Moreover, while the training must not necessarily be performed on edge devices, there is growing interest in federated learning~\cite{yang2019federated}, a privacy-preserving technology highly relevant to surveillance applications~\cite{truex2019hybrid}. 

We demonstrate a novel application of CycleGANs in the context of edge AI. Instead of collecting additional training data to extend the ODD of the smart camera, we used a CycleGAN to transform OOD input to the ODD. As CycleGANs learn style transformations from unpaired training data, this might enable a cost-efficient approach to ODD extension. In our case under study, we train a CyclaGAN to perform style transfers between the daytime and the nighttime domains.

In the underpass, we specified the ODD of the image recognition to perform classification in daytime conditions. In Experiment A, we trained a DNN model accordingly and report satisfactory results (cf. Table~\ref{tab:testset}). In Experiment B, we illustrate the limited robustness of the DNN model as it underperforms on OOD input, i.e., nighttime images. Subsequently, we used the CycleGAN to transform nighttime images to the daytime domain and reclassify the input. 

While the results are inconsistent across the different classes (cf. Figure~\ref{fig:cycleGANres}), we argue that the overall results indicate that the approach is promising: CycleGANs can make OOD input fit the ODD. Thanks to a learned style transfer, a small DNN model operating in tandem with a CycleGAN might be able to make predictions for input that goes beyond its training data, i.e., increasing the robustness of image recognition on constrained devices.

\section{Threats to Validity} \label{sec:threats}
All empirical research is subject to threats to validity. While the results reported in this study are preliminary, we report the main threats as our findings guide our future work on smart cameras for emergency management.

External validity reflects the generalizability of our results. Our initial work targets standard classes in image recognition, thus future work is needed to investigate how our findings extrapolate to classes customized for emergency management, e.g., person on the ground, a brawl, or bicycle accidents. Moreover, we did not study camera input with more than one class present at the same time. Future work should explore more complex activities in the underpass.

Furthermore, underpasses provide homogeneous environments for image recognition and our results cannot be extrapolated to less controlled public spaces such as pathways in parks. However, underpasses are prioritized locations for camera surveillance as they are known to be emergency hotspots. Finally, all video was recorded in the spring, thus we need to extend the dataset to cover seasonal variations.

Internal validity concerns casual relationships and potentially confounding factors. We report that neural network pruning and CycleGANs are promising approaches to enable efficient image recognition on edge devices. However, our conclusion is based on training DNNs using small datasets. It is possible that pruning would be less useful if the dataset was magnitudes larger. Moreover, perhaps a larger training set would also make the DNN robust enough to make CycleGAN transformations to the ODD superfluous.  

\section{Conclusion and Future Work} \label{sec:conc}
Edge AI paves the way for numerous applications of IoT for emergency management, e.g., image recognition in smart cameras. However, state-of-the-art DNN architectures are far from deployable on constrained edge devices. In this paper, we study DNN architectures for image classification for camera input in an underpass.

Our contributions are twofold. First, we report successful neural network pruning, i.e., we retain comparable classification accuracy using only 1.1\% of the size of the VGG16 architecture. Such a small DNN architecture can be deployed on constrained devices. Second, we propose that CycleGANs can be used to allow classification of OOD camera input by performing style transfers to the ODD. We present a proof-of-concept involving transforming nighttime input to the daytime domain, supporting the robustness of the application. The small DNN classifier can remain trained for only daytime conditions although the ODD of the image recognition solution can be extended to encompass additional environmental conditions. 

The preliminary work reported in this paper identified several interesting directions for future work. The obvious first step is to extend the dataset used for both the classification model and the CycleGAN. As the data labelling is labor-intensive, we plan to rely on our previous experience in active learning to focus annotation effort for maximum return on investment~\cite{borg2017using}. With more data, we can train the classifier to predict additional classes, including input related to emergency response.

Second, we plan to replace the image recognition with object recognition. While image recognition serves as a good first application, complementing the predictions with bounding boxes would be the natural next step. Accurate object detection and recognition can be made in real-time using DNN architectures such as YOLO~\cite{redmon2016you}. However, the prerequisite data labeling requires more manual effort.

Third, we intend to perform more systematic neural network pruning. In this paper, we explored the concept using an \textit{ad hoc} approach. However, state-of-the-art pruning involves sophisticated measurements of which neurons carry the most importance to the classification task~\cite{Molchanov2019ImportanceEF}.

\begin{figure}
\includegraphics[width=1.0\linewidth]{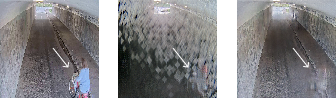}
\caption{Original daytime image (left), transformed to the spray domain (middle), and then reconstructed in the daytime domain (right). The arrows show the bicyclist.}
\label{fig:spray}
\end{figure}

Fourth, we will explore using CycleGANs as an approach to tackle vandalism. During the study, the camera dome in the underpass was targeted by an antagonistic spray paint attack. The image quality was compromised, but not totally useless. Figure~\ref{fig:spray} shows our initial efforts to learn a style transfer between the daytime domain and the sprayed domain. The results indicate that the approach deserves future study, and we propose that the concept could be used to temporarily recover from vandalism, especially for cameras mounted in difficult-to-reach locations.

\section{Acknowledgments}
This work was funded by Plattformen at Campus Helsingborg, Lund University.

\bibliographystyle{ACM-Reference-Format}
\bibliography{tunnel}

\end{document}